\documentclass[10pt,twocolumn,letterpaper]{article}

\usepackage{wacv}
\usepackage{times}
\usepackage{epsfig}
\usepackage{graphicx}
\usepackage{amsmath}
\usepackage{amssymb}
\usepackage{multirow}
\usepackage{nicefrac}
\usepackage{amsfonts}
\usepackage{amsthm}
\usepackage{booktabs}

\wacvfinalcopy 

\def\wacvPaperID{524} 

\ifwacvfinal\fi
\begin{document}

\title{Training Adversarial Discriminators for Cross-channel Abnormal Event Detection in Crowds}
\author{Mahdyar Ravanbakhsh$^{ 1, 3}$\\
		{\tt\small mahdyar.ravan@ginevra.dibe.unige.it}
		\and
		Enver Sangineto$^{ 2}$\\ \hspace{.5cm}
		{\tt\small enver.sangineto@unitn.it}
        \and
		\hspace{0.7cm}Moin Nabi$^{ 2, 3}$ \hspace{4.9cm}  Nicu Sebe$^{ 2}$\\
		\hspace{1.7cm}{\tt\small m.nabi@sap.com} \hspace{3.9cm} {\tt\small niculae.sebe@unitn.it}
		\and
		$^{1}$University of Genova, Italy   \hspace{.5cm} $^{2}$ University of Trento, Italy \hspace{.5cm} 
		$^{3}$ SAP SE., Berlin, Germany
}
\maketitle
\ifwacvfinal\thispagestyle{empty}\fi

\begin{abstract}
   Abnormal crowd behaviour detection attracts  a large interest due to its importance in video surveillance scenarios. However, the ambiguity and the lack of sufficient {\em abnormal} ground truth data makes end-to-end training of large deep networks hard in this domain.
In this paper we propose to use Generative Adversarial Nets (GANs), which are trained to generate 
only the {\em normal} distribution of the data. During  the adversarial GAN training, 
a discriminator ($D$) is used as a supervisor for the generator network ($G$) and vice versa.
At testing time we use $D$ to solve our  {\em discriminative} task (abnormality detection),  where $D$  has been trained without the need of manually-annotated  abnormal  data.
Moreover, in order to prevent $G$ learn a trivial identity function, we use a cross-channel approach, forcing $G$ to transform raw-pixel data in motion information and vice versa.
The quantitative results on standard benchmarks show 
that our method outperforms previous state-of-the-art methods in both the frame-level and the pixel-level evaluation.
\end{abstract}

\section{Introduction}
\label{sec:intro}

Detecting abnormal crowd behaviour is motivated by the increasing interest in video-surveillance systems for public safety. However, despite a lot of research has been done in this area in the past years \cite{li2014anomaly,kim2009observe,Mahadevan.anomaly.2010,mehran2009abnormal,lu2013abnormal,saligrama2012video,cong2011sparse,Feng,Nabi_2013_ICCV_Workshops,
	ravanbakhsh2016plug,sabokrouFFK16,7858798,sabokrou2016video,DBLP:conf/cvpr/0003CNRD16,xu2015learning,mousavi2015abnormality}, 
the problem is still open.


\begin{figure}
	\begin{minipage}[b]{0.9\linewidth}
		\centering
		\centerline{\includegraphics[width=\linewidth]{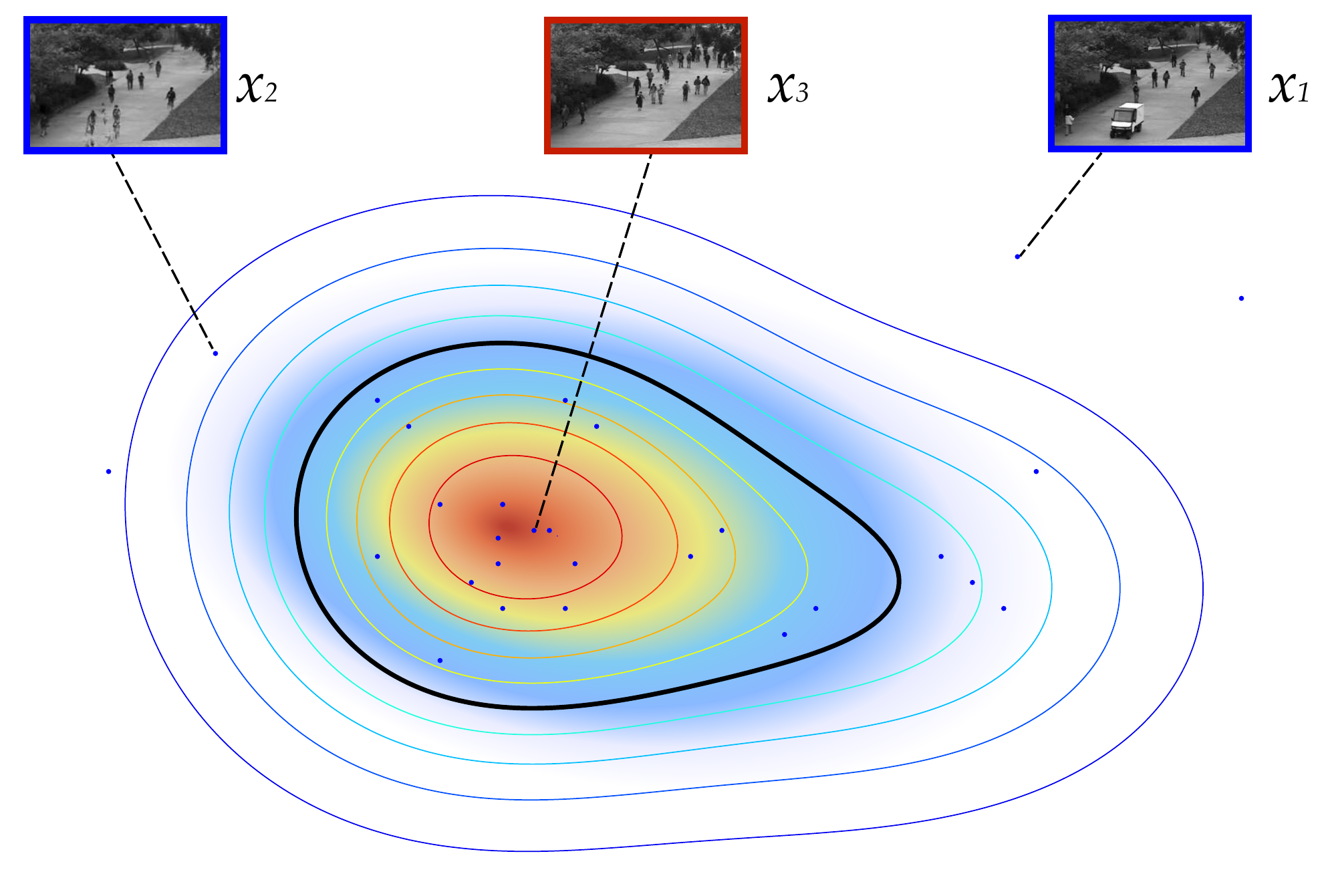}}
	\end{minipage}
	\caption{A schematic representation of our Adversarial Discriminator. The data distribution is denser in the feature space area corresponding to the only real and ``normal'' data observed by $G$ and $D$ during training. $D$ learns to separate this area from the rest of the feature space. In the figure, the solid black line represents the decision boundary learned by $D$. Outside this boundary lie both non-realistically generated images (e.g., $x_2$) and real but non-normal images (e.g., $x_1$). At testing time we exploit the learned decision boundary in order to detect abnormal events in new images.}
	\label{fig:teaser}
\end{figure}

One of the  main reasons for which abnormality detection is challenging is
the relatively small size of the
existing datasets with abnormality ground truth. 
In order to deal with this problem, 
most of the existing abnormality-detection methods focus on learning only the {\em normal} pattern of the crowd, 
for which only weakly annotated training data are necessary (e.g., videos representing only the normal crowd behaviour in a given scene).
Detection is then performed by comparing the the test-frame representation with  the previously learned normal pattern (e.g., using a one-class SVM~\cite{xu2015learning}). 


In this paper we propose to solve the abnormality detection problem using Generative Adversarial Networks (GANs) \cite{DBLP:conf/nips/GoodfellowPMXWOCB14}. GANs are deep networks mainly applied for unsupervised tasks and commonly used to generate data (e.g., images).
The supervisory information in a GAN is indirectly provided by an adversarial game between two independent networks: a generator ($G$) and a discriminator ($D$). 
During training, $G$ generates new data and $D$ tries to understand whether its input is real  (e.g., it is a training image) or it was generated by $G$. This competition between $G$ and $D$ is helpful in 
boosting the  accuracy of both $G$ and $D$.
At testing time, only $G$ is used to generate new data.

We use this framework to train our $G$ and $D$ using as training data only frames of videos without abnormality. Doing so, $G$ learns how to generate {\em only} the normal pattern of the observed scene. On the other hand, $D$ learns how to distinguish what is normal from what is not, because abnormal events are considered as outliers with respect to the data 
distribution  (see Fig.~\ref{fig:teaser}). Since our final goal is a discriminative task (at testing time we need to detect possible anomalies in a new scene), different from common GAN-based approaches, we propose to directly use $D$ after training.
The advantage of this approach is that we do not need to train one-class SVMs or other classifiers  on top of the learned visual representations and we present one of the very first deep learning approaches for abnormality detection which can be trained end-to-end.

As far as we know, the only other end-to-end deep learning framework for abnormality detection is the recently proposed approach of Hasan et al. \cite{DBLP:conf/cvpr/0003CNRD16}. In \cite{DBLP:conf/cvpr/0003CNRD16} a Convolutional 
Autoencoder is used to learn the crowd-behaviour normal pattern and used at testing time to {\em generate} the normal scene appearance, 
using the reconstruction error to measure an abnormality score.
The main difference of our approach with \cite{DBLP:conf/cvpr/0003CNRD16} is that we exploit the adversary game between $G$ and $D$ to simultaneously approximate the normal data distribution and train the final classifier.
In Sec.~\ref{sec:exp}-\ref{sec:ablation} we compare our method  with  both \cite{DBLP:conf/cvpr/0003CNRD16} and two strong baselines in which we use the reconstruction error of our generator $G$.
Similarly to \cite{DBLP:conf/cvpr/0003CNRD16}, in \cite{xu2015learning} Stacked Denoising Autoencoders  are used to reconstruct the input image and learn task-specific features using a deep network. However, in \cite{xu2015learning} the final classifier is a one-class SVM which is  trained on top of the learned representations and it is not jointly optimized together with the deep-network-based features. 

The second novelty we propose in this paper is 
a {\em multi-channel} data representation. Specifically, we use both appearance and motion (optical flow) information: a two-channel approach which has been proved to be empirically important in previous work on abnormality detection \cite{Mahadevan.anomaly.2010, ravanbakhsh2016plug,xu2015learning}.
Moreover, we propose to use a cross-channel approach where, inspired by \cite{Isola_2017_CVPR}, we train two networks which respectively transform raw-pixel images in optical-flow representations and vice versa.
The rationale behind this is that the architecture of  our conditional generators $G$ is  based on an encoder-decoder (see Sec.~\ref{sec:tasks}) 
and we use these channel-transformation tasks in order  
to prevent $G$ learn a trivial identity function and force 
$G$ and $D$  to construct sufficiently informative internal representations.



In the rest of this paper we  review the related literature in Sec.~\ref{sec:RelatedWork} and we present our method in Sec.~\ref{sec:tasks}-\ref{sec:Detection}. Experimental results are reported in Sec.~\ref{sec:exp}-\ref{sec:ablation}. Finally, we show some qualitative results in Sec.~\ref{sec:qualitative} and we conclude in Sec.~\ref{sec:Conclusions}.

\section{Related Work}
\label{sec:RelatedWork}

In this section we briefly review previous work considering: (1) our application scenario (Abnormality Detection) and (2) our methodology based on GANs.

{\bf Abnormality Detection} There is a wealth of literature on abnormality detection~\cite{rabiee2017detection,li2014anomaly,mehran2009abnormal,sebe2018abnormal,rabiee2016novel,mousavi2015analyzing,Mahadevan.anomaly.2010,cong2011sparse,kim2009observe,saligrama2012video,lu2013abnormal,rabiee2016emotion,rabiee2016crowd,ravanbakhshfast}. Most of the previous work is based on  hand-crafted features (e.g., Optical-Flow, Tracklets, etc.) to model the normal activity patterns, whereas our method learns features from raw-pixels using a deep-learning based approach using an end-to-end training protocol. 
Deep learning has also been investigated for abnormality detection tasks in~\cite{ravanbakhsh2016plug,sabokrouFFK16,7858798}. 
Nevertheless, these works mainly use existing Convolutional Neural Network (CNN) models trained for other tasks (e.g., object recognition) which are adapted to the abnormality detection task. For instance, Ravanbakhsh et al.~\cite{ravanbakhsh2016plug} proposed a Binary Quantization Layer, plugged as a final layer on top of a pre-trained CNN, in order to represent patch-based   temporal motion patterns. However, the network proposed in \cite{ravanbakhsh2016plug} is not trained end-to-end and is based on a complex post-processing stage and on a pre-computed codebook of the convolutional feature  values. Similarly, in \cite{sabokrouFFK16,7858798}, a fully convolutional neural network is proposed which is a combination of  a pre-trained CNN (i.e., AlexNet \cite{alexnet}) and a new convolutional layer where kernels have been trained from scratch. 

Stacked Denoising Autoencoders (SDAs) are used by Xu et al.~\cite{xu2015learning} to learn motion and appearance feature representations. The networks used in this work are relatively shallow, since training deep SDAs on small abnormality datasets can be prone to  over-fitting issues and the networks' input is limited to a small image patch.
Moreover, after the SDAs-based features have been learned, multiple one-class SVMs need to be trained on top of these features in order to create the final classifiers, and the learned features may be sub-optimal because they are not jointly optimized with respect to the final abnormality discrimination task.
Feng et al. \cite{Feng} use  3D gradients and a PCANet \cite{PCANet} in order to extract patch-based appearance features whose normal distribution is then modeled using a deep Gaussian Mixture Model  network (deep GMM \cite{deepGMM}). Also in this case the feature extraction process and the normal event modeling are obtained using two separate stages (corresponding to two different networks) and the lack of an end-to-end training which jointly optimizes  both these stages can likely produce sub-optimal representations. Furthermore, the number of Gaussian components in each layer of the deep GMM is a critical hyperparameter which needs to be set using supervised validation data.

The only deep learning based approach proposing a framework which can be  fully-trained in an end-to-end fashion we are aware of is the Convolutional AE network proposed in \cite{DBLP:conf/cvpr/0003CNRD16}, where a deep representation is learned by minimizing the AE-based frame reconstruction. At testing time, an anomaly is detected computing the difference between the AE-based frame reconstruction and the real test frame.
We compare with this work 
in Sec.~\ref{sec:exp} and in  Sec.~\ref{sec:ablation} we present two modified versions of our GAN-based approach ({\em Adversarial Generator} and {\em GAN-CNN}) in which, similarly to \cite{DBLP:conf/cvpr/0003CNRD16}, we use the reconstruction errors of our adversarially-trained generators as detection strategy.
Very recently, Ravanbakhsh et al.~\cite{2017icip} proposed to use the reconstruction errors of the generator networks to detect anomalies at testing time instead of directly using the corresponding discriminators as we propose here. However, their method needs  an externally-trained CNN to capture sufficient semantic information and  a fusion strategy which takes into account  the reconstruction errors of the two-channel generators. Conversely, the discriminator-version proposed in this paper is simpler to reproduce and faster to run. Comparison between these two versions is provided in Sec.~\ref{sec:ablation}, together with a detailed ablation study of all the elements of our proposal.
\noindent
{\bf GANs} \cite{DBLP:conf/nips/GoodfellowPMXWOCB14,DBLP:conf/nips/SalimansGZCRCC16,DBLP:journals/corr/RadfordMC15,Isola_2017_CVPR,nguyen2016ppgn} are based on a two-player game between two different networks, both trained with unsupervised data. One network is the {\em generator} ($G$), which aims at generating realistic data (e.g., images). The second network is the {\em discriminator} ($D$), which aims at discriminating real data from data generated from $G$. 
Specifically, the {\em conditional} GANs \cite{DBLP:conf/nips/GoodfellowPMXWOCB14}, that we use in our approach,
are trained with a set of data point pairs (with loss of generality, from now on we assume both data points are images): $\{ (x_i, y_i) \}_{i=1,...,N}$, where image $x_i$ and image $y_i$ are somehow each other semantically related.
$G$ takes as input $x_i$ and  random noise $z$  and generates a new image $r_i = G(x_i, z)$. 
$D$ tries to distinguish $y_i$ from $r_i$, while $G$ tries to ``fool'' $D$ producing more and more realistic images which are hard to be distinguished.\\
Very recently Isola et al. \cite{Isola_2017_CVPR} proposed an ``image-to-image translation'' framework based on conditional GANs, where both the generator and the discriminator are conditioned on the real data. They show that a ``U-Net'' encoder-decoder with skip connections can be used as the generator architecture together with a patch-based discriminator in order to transform images with respect to different representations.
We adopt this framework in order to generate optical-flow images from raw-pixel frames and vice versa. However, it is worth to highlight that, different from common GAN-based approaches, we do not aim at generating image representations which look realistic, but we use $G$ to learn the normal pattern of an observed crowd scene. At testing time, $D$ is directly used to detect abnormal areas using the 
appearance and the motion of the input frame.

\begin{figure}
	\begin{minipage}[b]{0.99\linewidth}
		\centering
		\centerline{\includegraphics[width=\linewidth]{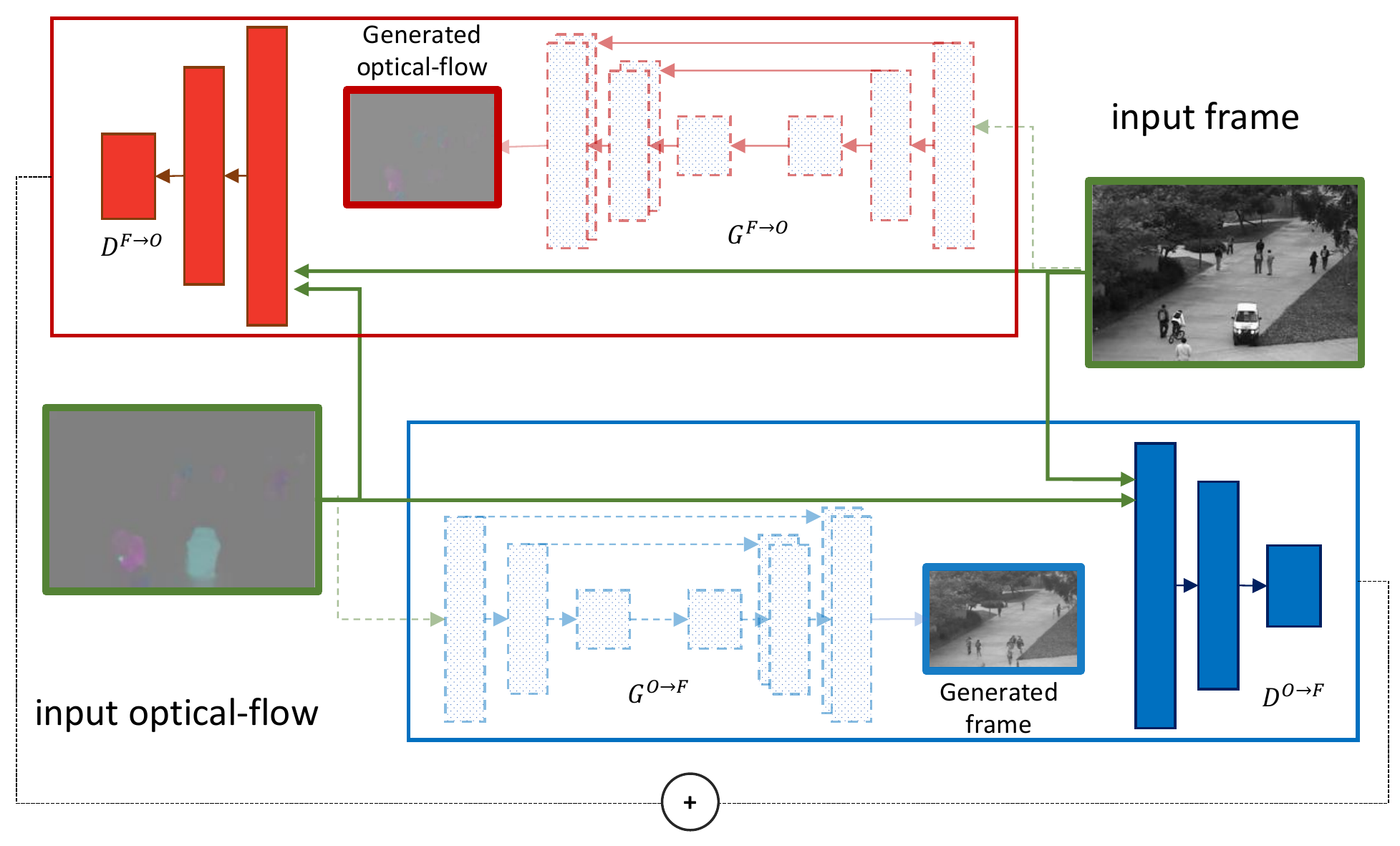}}
	\end{minipage}
	\caption{A schematic representation of our proposed detection method. }
	\label{fig:over_temp}
\end{figure}

\section{Cross-channel Generation Tasks}
\label{sec:tasks}
Inspired by Isola et al. \cite{Isola_2017_CVPR}, we built our freamwork to learn the normal behaviour of the crowd in the observed scene. 
We use two channels: appearance (i.e., raw-pixels) and motion (optical flow images) and two cross-channel tasks. In the first task, we generate optical-flow images starting from the original frames, while in the second task we generate appearance information starting from an optical flow image.

Specifically, let $F_t$ be the $t$-th frame of a training video and $O_t$ the optical flow obtained using $F_t$ and $F_{t+1}$. $O_t$ is computed using \cite{brox2004high}.
We train two networks: ${\cal N}^{F \rightarrow O}$, which generates optical-flow from frames (task 1) and 
${\cal N}^{O \rightarrow F}$, which generates frames from optical-flow (task 2).
In both cases, our networks are composed of a conditional generator $G$
and a conditional discriminator $D$. $G$ takes as input an image $x$ and a noise vector $z$ (drawn from a noise distribution ${\cal Z}$) and outputs an image $r = G(x,z)$ 
of the same dimensions of $x$ 
but represented in a different channel. 
For instance, in case of ${\cal N}^{F \rightarrow O}$, $x$ is a frame ($x  = F_t$) and $r$ is {\em the reconstruction} of its corresponding optical-flow image $y  = O_t$. On the other hand,
$D$ takes as input two images: $x$ and $u$ (where $u$ is either $y$ or $r$) and outputs a scalar representing the probability that both its input images came from the real data.

Both $G$ and $D$ are fully-convolutional networks, composed of convolutional layers, batch-normalization layers and ReLU nonlinearities.
In case of $G$ we adopt the U-Net architecture~\cite{ronneberger2015u}, which is  an encoder-decoder, where the input $x$ is passed through a series of progressively downsampling layers until a bottleneck layer, at which point the forwarded information is upsampled. Downsampling and upsampling layers in a symmetric position with respect to the bottleneck layer are connected by {\em skip connections} which help preserving important local information. The noise vector $z$ is implicitly provided to $G$ using dropout, applied to multiple layers.

The two input images $x$ and $u$ of $D$ are concatenated and passed through 5 convolutional layers. In more detail, $F_t$ is represented using the standard RGB representation, while 
$O_t$ is represented using the horizontal, the vertical and the magnitude components. Thus, in both tasks, the input of $D$ is composed of 6 components (i.e., 6 2D images), whose relative order depends on the specific task. 
All the images are rescaled to $256 \times 256$.
We use the popular {\em PatchGAN} discriminator~\cite{li2016precomputed}, which is
based on a ``small'' fully-convolutional discriminator $\hat{D}$.
$\hat{D}$ is
applied to a $30 \times 30$ grid, where each position of the grid corresponds to a $70 \times 70$ patch $p_x$ in $x$ and a corresponding patch $p_u$ in $u$. The output of $\hat{D}(p_x,p_u)$ is a score representing the probability 
that $p_x$ and $p_u$ are  both real. 
During training, the output of $\hat{D}$ over all the grid positions  is averaged
and this provides the final score of $D$ with respect to $x$ and $u$.
Conversely, at testing time we directly use $\hat{D}$ as a ``detector'' which is run over the grid to spatially localize the possible abnormal regions in the input frame (see Sec.~\ref{sec:Detection}).

\section{Training}
\label{sec:training}

$G$ and $D$ are trained using both a conditional GAN loss ${\cal L}_{cGAN}$ and a reconstruction loss
${\cal L}_{L1}$. In case of ${\cal N}^{F \rightarrow O}$, the training set is composed of pairs of frame-optical flow images
${\cal X} = \{ (F_t, O_t) \}_{t=1,...,N}$. 
${\cal L}_{L1}$ is given by:
\begin{equation}
{\cal L}_{L1}(x,y) =  ||y - G(x,z) ||_1,
\end{equation}
\noindent
where $x = F_t$ and $y = O_t$,
while the conditional adversarial loss ${\cal L}_{cGAN}$ is:
\begin{align}
{\cal L}_{cGAN}(G,D)= 
\mathbb{E}_{(x,y) \in {\cal X}} [\log D(x,y)] + \\
\mathbb{E}_{x \in \{ F_t \}, z \in {\cal Z}} [\log ( 1 - D(x,G(x,z)) )]
\end{align}
Conversely, in case of ${\cal N}^{O \rightarrow F}$, we use ${\cal X} = \{ (O_t, F_t) \}_{t=1,...,N}$. 
What is important to highlight here is that both $ \{ F_t \}$ and $\{ O_t \}$ are collected 
using the frames of the only {\em normal} videos of the training dataset. 
The fact that we do not need videos showing abnormal events at training time makes it possible to train the  discriminators  corresponding to our two  tasks without the need of supervised training data: $G$ acts as an implicit supervision for $D$ (and vice versa).


%
%

\begin{figure*}
\begin{tabular}{ c c c }
        I - Images generated by $G^{O \rightarrow F}$  & & II - Optical flow generated by $G^{F \rightarrow O}$ \\
		\scriptsize{(a)}
		\includegraphics[width=.23\linewidth]{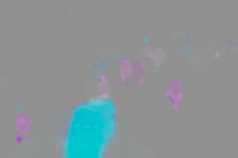}
		\includegraphics[width=.23\linewidth]{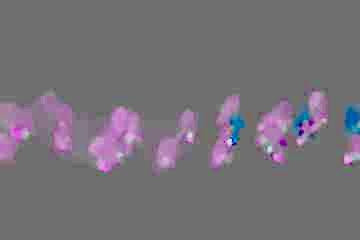} && 
		\scriptsize{(a)}  
		\includegraphics[width=.23\linewidth]{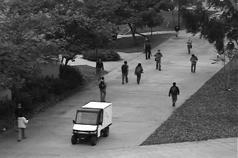} 
		\includegraphics[width=.23\linewidth]{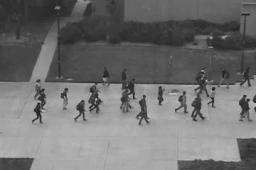} 
		\\
		
		\scriptsize{(b)}
		\includegraphics[width=.23\linewidth]{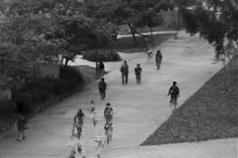}
		\includegraphics[width=.23\linewidth]{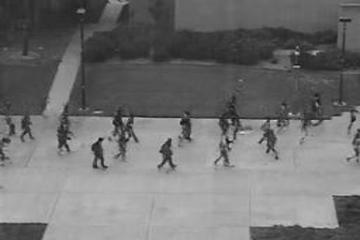}&& \scriptsize{(b)}
		\includegraphics[width=.23\linewidth]{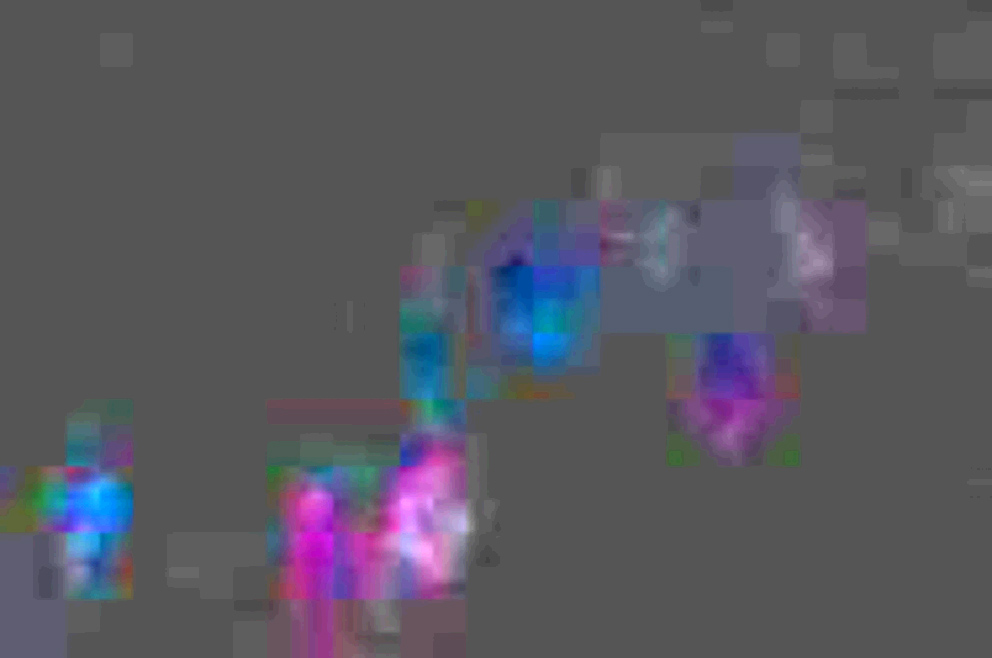}
		\includegraphics[width=.23\linewidth]{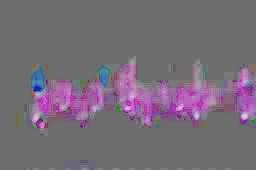}
		\\
		
		\scriptsize{(c)}
		\includegraphics[width=.23\linewidth]{images/ped1_1}
		\includegraphics[width=.23\linewidth]{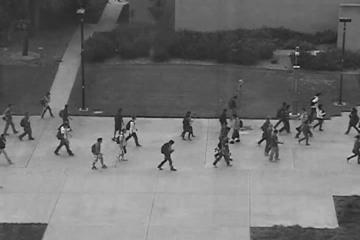}&&
		\scriptsize{(c)}
		\includegraphics[width=.23\linewidth]{images/ped1_5}
		\includegraphics[width=.23\linewidth]{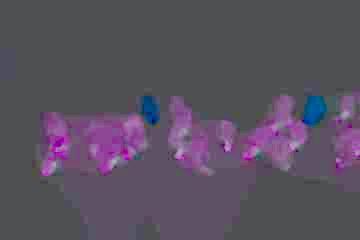}
		\\
	\end{tabular}
	\caption{ A  few  examples  of  generations after training is completed: \textbf{(I) Images generated by $G^{O \rightarrow F}$}: (a) the input optical-flow images, (b) the corresponding generated  frames, (c) the real frames corresponding to (a). \textbf{(II) Optical flow images generated by $G^{F \rightarrow O}$}: (a) the real input frames, (b) the corresponding generated optical flow images, (c) the real optical flow images corresponding to (a). The first column represent an abnormal scene, while the other column depicts a normal situation. Note that the source of abnormality (the vehicle) in both cases has not been reconstructed correctly.}
	\label{fig:qual2-O-F}
\end{figure*}

During training the  generators of the two tasks ($G^{F \rightarrow O}$ and $G^{O \rightarrow F}$) observe only normal scenes. As a consequence, after training they are not able to reconstruct an abnormal event. 
For instance, in Fig.~\ref{fig:qual2-O-F} (II) a frame $F$ containing a vehicle unusually moving in a University campus
is input to $G^{F \rightarrow O}$ and in the generated optical flow image ($r_O = G^{F \rightarrow O}(F)$) the abnormal area corresponding to that vehicle is not properly reconstructed. Similarly, when the real optical flow ($O$) associated with $F$ is input to $G^{O \rightarrow F}$, the network tries to reconstruct the area corresponding to the vehicle but the output is a set of unstructured blobs (Fig.~\ref{fig:qual2-O-F}, first column).
On the other hand, the two corresponding discriminators $D^{F \rightarrow O}$ and $D^{O \rightarrow F}$ during training have learned to distinguish what is plausibly 
real in the given scenario from what is not and we will exploit this learned discrimination capacity at testing time.

Note that, even if a global optimum can be theoretically reached in a GAN-based training, in which 
the  data distribution and the generative distribution totally overlap each other
\cite{DBLP:conf/nips/GoodfellowPMXWOCB14}, in practice the generator is very rarely able to generate fully-realistic images. For instance, in Fig.~\ref{fig:qual2-O-F} the high-resolution details of the generated pedestrians (``normal'' objects)  are quite smooth and the human body is approximated with a blob-like structure. As a consequence, at the end of the training process, the discriminator has learned to separate real data from artifacts.
This situation is schematically represented in Fig.~\ref{fig:qual2-O-F}. 
 The discriminator is represented by the decision boundary on the learned feature space which separates the densest area of this distribution from the rest of the space. Outside this area lie both non-realistic generated images (e.g. $x_2$) and real, abnormal events (e.g., $x_1$).
Our hypothesis is that the latter lie outside the discriminator's decision boundaries because they represent situations never observed during training and hence treated by $D$ as outliers.
We use the discriminator's learned decision boundaries in order to detect $x_1$-like events as explained in the next section.

\section{Abnormality Detection}
\label{sec:Detection}

At testing time only the discriminators are used. More specifically, 
let $\hat{D}^{F \rightarrow O}$ and $\hat{D}^{O \rightarrow F}$ be the patch-based discriminators trained using the two channel-transformation tasks (see Sec.~\ref{sec:tasks}). 
Given a test frame $F$ and its corresponding optical-flow image $O$, we apply the two patch-based discriminators on the same $30 \times 30$ grid used for training. This results in two $30 \times 30$ score maps: $S^O$ and $S^F$ for the first and the second task, respectively. 
Note that we do not need to produce the reconstruction images to use the discriminators. For instance, for a given position on the grid, $\hat{D}^{F \rightarrow O}$ takes as input a patch 
$p_F$ on $F$ and a corresponding patch 
$p_O$ on $O$. 
A possible abnormal area in $p_F$ and/or in $p_O$ (e.g., an unusual object or an unusual movement) corresponds to an outlier with respect to the  distribution learned by $\hat{D}^{F \rightarrow O}$ during training and results in a low value of  $\hat{D}^{F \rightarrow O}(p_F,p_O)$. By setting a threshold on this value we obtain a decision boundary (see Fig.~\ref{fig:teaser}). However, following a common practice, we first fuse the channel-specific score maps and then we apply a range of confidence thresholds on the final abnormality map in order to obtain different ROC points (see Fig \ref{fig:over_temp} and Sec.~\ref{sec:exp}). Below we show how the final abnormality map is constructed.

The two score maps are summed with equal weights: $S= S^O + S^F$.
The values in $S$ are normalized in the range $[0, 1]$. In more detail, for each test video $V$
we compute the maximum value $m_s$ of all the elements of $S$ over all the input frames of $V$. 
For each frame the normalized  score map is given by:
\begin{equation}
N(i,j) = 1/m_s S(i,j), i,j \in \{1, ..., 30\}
\end{equation}
Finally, we upsample $N$  to the original frame size 
($N'$)
and the previously computed optical-flow is used to filter out non-motion areas, obtaining the final abnormality map:
\begin{equation}
\label{eq.filtering}
A(i,j) = \left\{%
\begin{array}{ll}
1 - N'(i,j)  & \mbox{if $O(i,j) > 0$}   \\
0  & \mbox{otherwise.}    \\
\end{array}
\right.
\end{equation}
Note that all the post-processing steps (upsampling, normalization, motion-based filtering) are quite common strategies for abnormal-detection systems \cite{xu2015learning} and we do not use any hyper-parameter or ad-hoc heuristic which need to be tuned on a specific dataset.


\section{Experimental Results} 
\label{sec:exp}



In this section we compare the proposed method
against the state of the art   
using  common benchmarks for crowd-behaviour abnormality detection. The evaluation is performed using both a {\em pixel-level} and a {\em frame-level} protocol and the  evaluation setup proposed in \cite{li2014anomaly}. The rest of this section describes the datasets, the experimental protocols and the obtained results.


\begin{figure*}
	\begin{minipage}[b]{.5\linewidth}
		\centering
		\centerline{\includegraphics[width=0.85\textwidth]{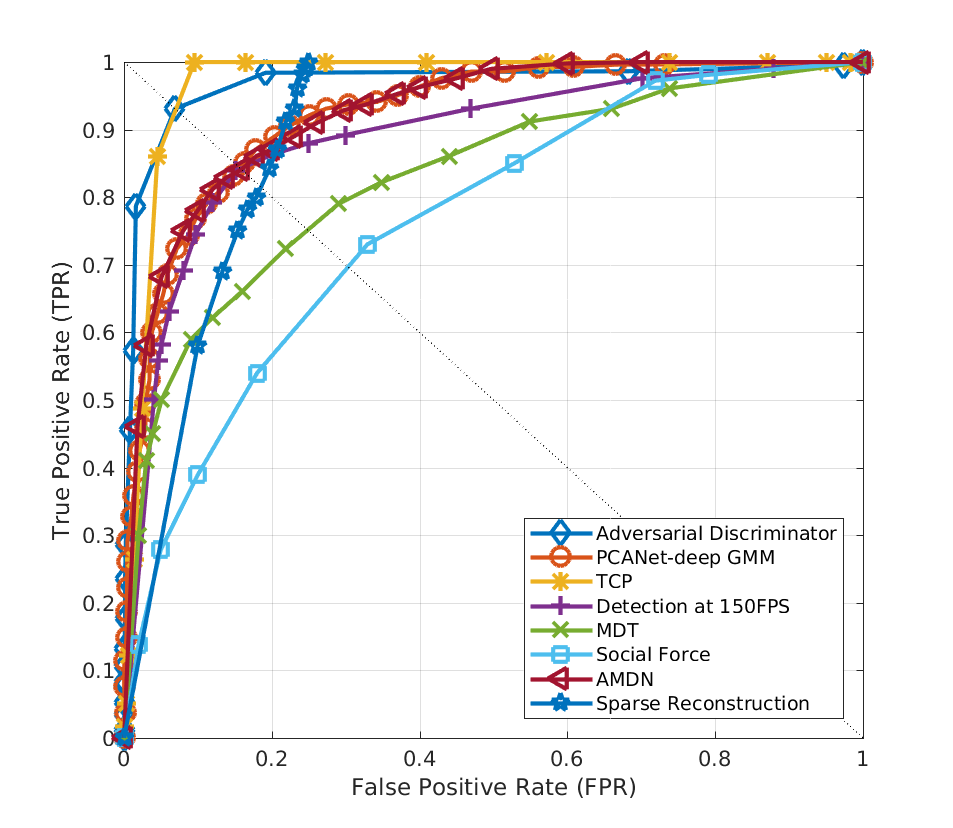}}
		\centerline{\scriptsize(a) frame-level ROC}\medskip
	\end{minipage}
	\begin{minipage}[b]{0.5\linewidth}
		\centering
		\centerline{\includegraphics[width=0.85\textwidth]{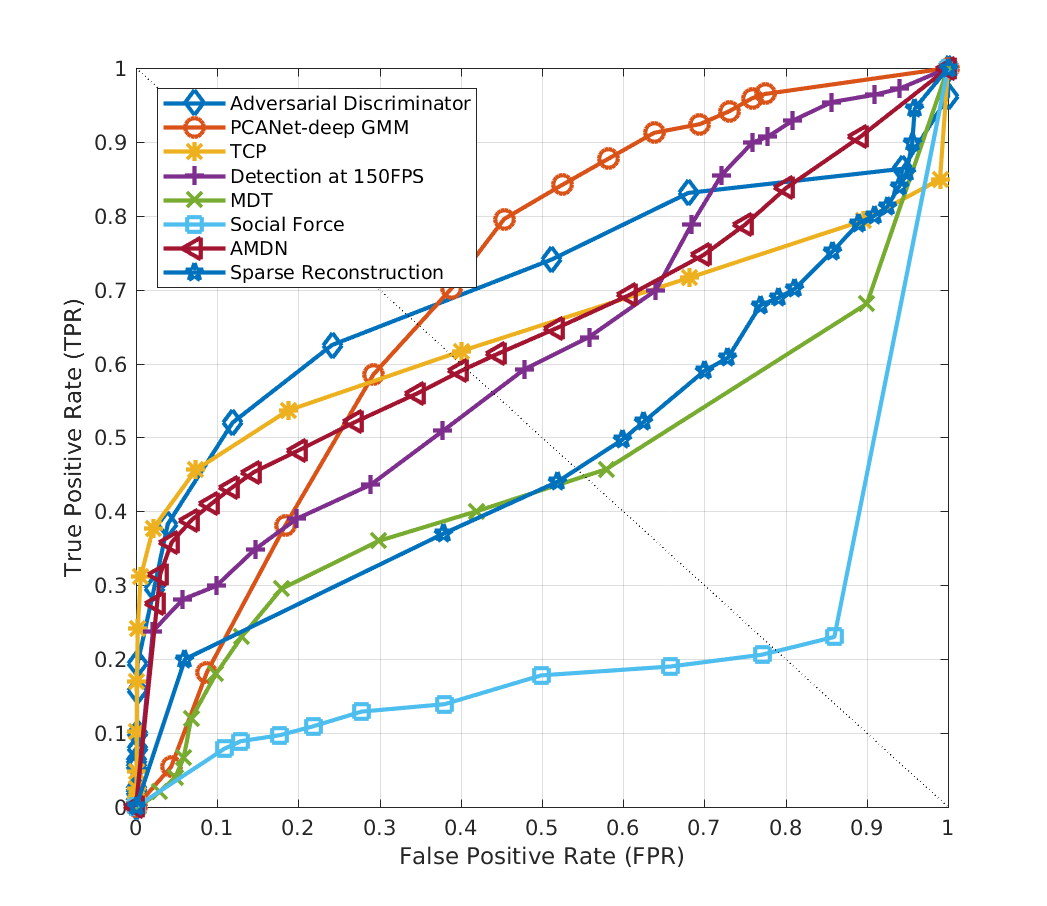}}
		\centerline{\scriptsize(b) pixel-level ROC}\medskip
	\end{minipage}
	\caption{ROC curves for Ped1 (UCSD dataset).}
	\label{fig:rocfrm}
\end{figure*}


\begin{table*}
	\begin{center}
		\resizebox{0.9\textwidth}{!}{
		\begin{tabular}[width=\textwidth]{l cc l cc l cc}
			\toprule
			\multirow{2}{*}{Method} &\multicolumn{2}{c}{Ped1 (frame-level)} & &\multicolumn{2}{c}{Ped1 (pixel-level)} & &\multicolumn{2}{c}{Ped2 (frame-level)}\\ \cmidrule{2-3} \cmidrule{5-6} \cmidrule{8-9}
			& EER & AUC & & EER & AUC && EER & AUC\\
			\midrule
			MPPCA~\cite{kim2009observe} & 		            40\%& 	59.0\% & 	&	81\%& 20.5\%& 	&	    30\%& 69.3\%\\
			Social force (SFM)~\cite{mehran2009abnormal} &    31\% & 	 67.5\% & 	&	79\%& 19.7\%& 	&	    42\%& 55.6\%\\
			SF+MPPCA~\cite{Mahadevan.anomaly.2010} & 		32\% & 	 68.8\%& 	&	71\%& 21.3\%& 	&		36\%& 61.3\%\\
			Sparse Reconstruction~\cite{cong2011sparse} & 		                19\%&   ---     & 	&	54\%& 45.3\%& 	&	    --- & --- \\
			MDT~\cite{Mahadevan.anomaly.2010} & 		    25\% &  81.8\% & 	&	58\%& 44.1\%& 	&		25\%& 82.9\%\\
			Detection at 150fps~\cite{lu2013abnormal} & 	15\%& 	91.8\% & 	&	43\%& 63.8\%& 	&		--- & ---\\
			TCP~\cite{ravanbakhsh2016plug} &  8\% &   95.7\%&     &   40.8\%& 64.5\%& &       18\%& 88.4\%\\
			AMDN (double fusion)~\cite{xu2015learning} & 	16\% & 	 92.1\%& 	&	40.1\%& 67.2\%& &		17\%& 90.8\%\\
			Convolutional AE~\cite{DBLP:conf/cvpr/0003CNRD16} & 	27.9\% & 	 81\%& 	&	--- & --- & &		21.7\% & 90\%\\
			PCANet-deep GMM~\cite{Feng} & 	15.1\% & 	 92.5\%& 	&	35.1\% & 69.9\% & &		--- & ---\\
			Adversarial Discriminator &               \textbf{7\%} & \textbf{96.8\%} &   &\textbf{34\%}& \textbf{70.8\%}& &\textbf{11\%}& \textbf{95.5\%}\\
			\bottomrule
		\end{tabular}
		}
	\end{center}
	\caption{UCSD dataset. Comparison of different methods. The results of {\em PCANet-deep GMM} are taken from \cite{Feng}. The other results but ours are taken  from~\cite{xu2015learning}.}
	\label{tbl:UCSD}
\end{table*}
\noindent\textbf{Implementation details.}
${\cal N}^{F \rightarrow O}$ and ${\cal N}^{O \rightarrow F}$ are trained using  the training sequences of the UCSD dataset (containing only ``normal'' events). 
All frames are resized to $256 \times 256$ pixels (see Sec.~\ref{sec:tasks}). Training is based on stochastic gradient descent with momentum 0.5 and batch size 1. 
We train our networks for 10 epochs each.
All the  GAN-specific hyper-parameter values have been set following the suggestions in \cite{Isola_2017_CVPR}, 
while in our approach there is no dataset-specific hyper-parameter which needs to be tuned. This makes the proposed method particularly robust, especially in a weakly-supervised scenario in which ground-truth validation data with abnormal frames are not given. 
All the results presented in this section but ours are taken from \cite{xu2015learning,mousavi2015crowd} 
which report the best results  achieved by each method independently tuning the method-specific hyper-parameter values.

Full-training of one network (10 epochs) takes on average less than half an hour with 6,800 training samples.
At testing time, one frame is processed in 0.53 seconds (the whole processing pipeline, optical-flow computation and post-processing included). These computational times have been computed using a single GPU (Tesla K40).

\noindent\textbf{Datasets and experimental setup.}
We use two standard datasets: the UCSD Anomaly Detection Dataset~\cite{Mahadevan.anomaly.2010} and the UMN SocialForce~\cite{mehran2009abnormal}. The \textbf{UCSD dataset} is split into two subsets: {\em Ped1}, which is composed of 34 training and 16 test sequences, and {\em Ped2}, which is composed of 16 training and 12 test video samples.
The overall dataset 
contains about  3,400 abnormal and 5,500 normal frames. 
This dataset is challenging due to the low resolution of the images and the presence of different types of moving objects and anomalies in the scene. The \textbf{UMN dataset} contains 11 video sequences in 3 different scenes, with a total amount of 7,700 frames. All the sequences start with a normal frame and end with an abnormal frame.\\
\noindent\textbf{Frame-level evaluation:}
In the frame-level anomaly detection evaluation protocol,
an abnormality label is predicted for a given test frame if at least  one abnormal pixel is predicted in that frame: In this case the abnormality label is assigned to the whole frame. This evaluation procedure is iterated using a range of confidence thresholds  in order to build a corresponding ROC curve. 
In our case, these confidence thresholds are directly applied to the output of the abnormality map $A$ defined in Eq.~\ref{eq.filtering} (see Sec.~\ref{sec:Detection}).
The results are reported in Tab.~\ref{tbl:UCSD} (UCSD dataset)
and Tab.~\ref{tbl:umn} (UMN dataset)
using the Equal Error Rate (EER) and the Area Under Curve (AUC).
Our method is called {\em Adversarial Discriminator}.
Fig.~\ref{fig:rocfrm} (a)
shows  the ROC curves (UCSD dataset).\\
\begin{table}
	\begin{center}
		\begin{tabular}[width=\textwidth]{l c c}
			\toprule
			Method 							&			& 	AUC \\
			\midrule
			Optical-flow~\cite{mehran2009abnormal}	  &  &	0.84 \\
			Social force (SFM)~\cite{mehran2009abnormal} 		&		&	0.96\\
			Sparse Reconstruction~\cite{cong2011sparse} & &  0.97\\
			Commotion Measure~\cite{mousavi2015crowd} 	&	    &	0.98\\
			TCP~\cite{ravanbakhsh2016plug}&&	0.98\\
			Adversarial Discriminator 		&					&	\textbf{0.99}\\
			\bottomrule
			
		\end{tabular}
	\end{center}
	\caption{UMN dataset. Comparison of different methods. All but our results are taken  from~\cite{mousavi2015crowd}.}
	\label{tbl:umn}
\end{table}
\noindent\textbf{Pixel-level anomaly localization:}
The goal of the pixel-level evaluation is to measure the accuracy of the abnormality spatial {\em localization}. Following the protocol suggested in~\cite{li2014anomaly}, the predicted abnormal pixels are compared with the pixel-level ground truth. 
A test frame is a true positive if the area of the predicted abnormal pixels overlaps with the ground-truth area by 
at least $40\%$, 
otherwise the frame is counted as a false positive. 
Fig.~\ref{fig:rocfrm} (b) shows the ROC curves of the localization accuracy over the USDC dataset, and EER and AUC
values are reported in Tab.~\ref{tbl:UCSD}.



\begin{figure*}
	\begin{center}
		\scriptsize{(a)}
		\includegraphics[width=.23\linewidth]{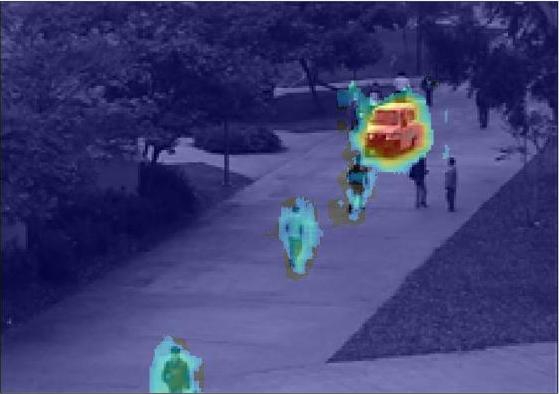}
		\includegraphics[width=.23\linewidth]{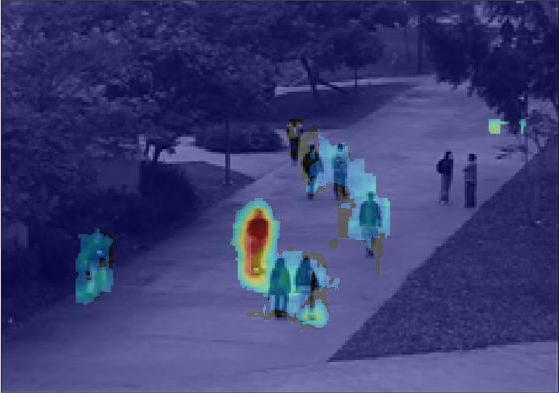}
		\includegraphics[width=.23\linewidth]{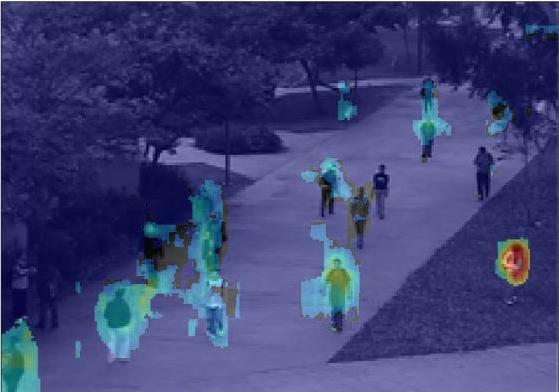}
		\includegraphics[width=.242\linewidth]{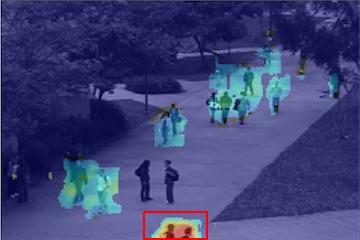}
		
		\scriptsize{(b)}
		\includegraphics[width=.23\linewidth]{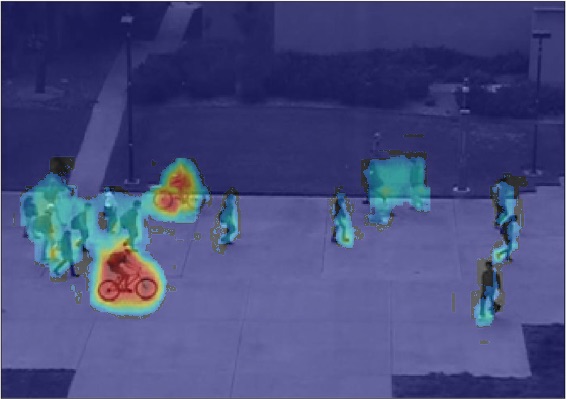}
		\includegraphics[width=.23\linewidth]{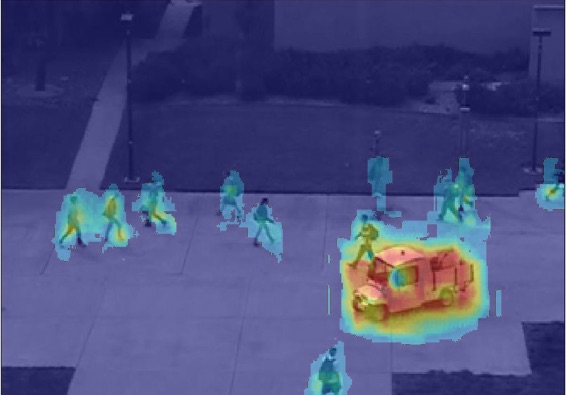}
		\includegraphics[width=.23\linewidth]{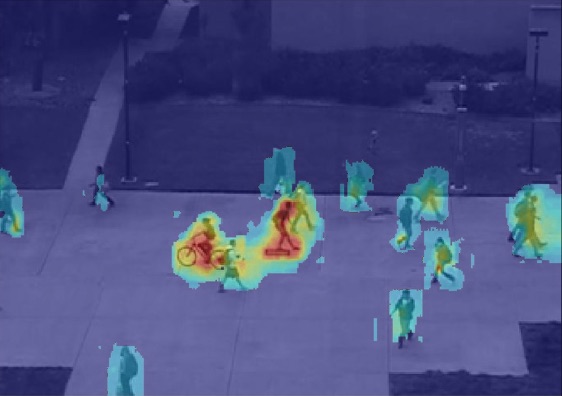}
		\includegraphics[width=.242\linewidth]{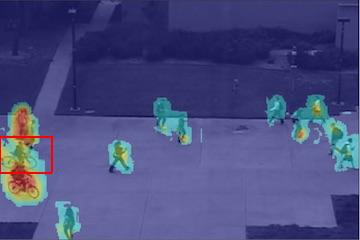}
		
	\end{center}
	\caption{A few examples of pixel-level detections of our method, visualizing the abnormality score using heat-maps. (a) Ped1 dataset, (b) Ped2 dataset. The last column shows some examples of detection errors of our method. The red rectangles highlight the prediction errors.}
	\label{fig:qual-det-succ}
\end{figure*}
\section{Ablation Study}
\label{sec:ablation}

In this section we analyse the main aspects of the proposed method, which are: the use of the discriminators trained by our conditional GANs as the final classifiers, the importance of the cross-channel tasks and the influence of the multiple-channel approach (i.e., the importance of fusing appearance and motion information).
For this purpose we use the UCSD Ped2  dataset (frame-level evaluation)
and we test different strong baselines obtained by amputating important aspects of our method.

The first baseline, called {\em Adversarial Generator}, is obtained using the reconstruction error of $G^{F \rightarrow O}$ and $G^{O \rightarrow F}$, which are the  generators trained as in Sec.~\ref{sec:tasks}-\ref{sec:training}. In more detail,  at testing time we use $G^{F \rightarrow O}$ and $G^{O \rightarrow F}$ to {\em generate} a channel transformation of the input frame $F$ and its corresponding optical-flow image $O$. Let $r_O = G^{F \rightarrow O}(F)$ and $r_F = G^{O \rightarrow F}(O)$. Then, similarly to Hasan et al. \cite{DBLP:conf/cvpr/0003CNRD16}, we compute the appearance reconstruction error using: $e_F = |F - r_F|$ and the motion reconstruction error using: $e_O = |O - r_O|$. When an anomaly is present in $F$ and/or in $O$, $G^{F \rightarrow O}$ and $G^{O \rightarrow F}$ are not able to accurately reconstruct the corresponding area (see Sec.~\ref{sec:qualitative} and Fig.~\ref{fig:qual2-O-F}). Hence, we expect that, in correspondence with  these abnormal areas, $e_F$ and/or $e_O$ have higher values than 
the average values computed when using normal test frames. 
The final abnormality map is obtained by applying the same post-processing steps described in Sec.~\ref{sec:Detection}:
(1) we upsample the reconstruction errors, 
(2) we normalize the
the two  errors with respect to all the  frames in the test video $V$ and in each channel independently of the other channel,
(3) we fuse the normalized maps and (4) we use optical-flow to filter-out non-motion areas. The only difference with respect to the corresponding post-processing stages adopted in case of {\em Adversarial Discriminator} and described in Sec.~\ref{sec:Detection} is a weighted fusion of the channel-dependent maps by weighting the importance of $e_O$ twice as the importance of $e_F$. 

\begin{table}
	\begin{center}
		\resizebox{0.4\textwidth}{!}{
		\begin{tabular}[width=\textwidth]{l cc}
			\toprule
			Baseline &  EER & AUC \\
			\midrule
			Adversarial Generator & 15.6\% & 93.4\%\\
			Adversarial Discriminator F & 24.9\%& 81.6\%\\
			Adversarial Discriminator O &13.2\% &90.1\% \\
			Adversarial Discriminator &\textbf{11\%}& \textbf{95.5\%}\\
			GAN-CNN &\textbf{11\%}& 95.3\%\\
			\toprule
		\end{tabular}
		}
	\end{center}
	\caption{Results of the ablation analysis on the UCSD dataset, Ped2 (frame-level evaluation).}
	\label{tbl:ablation}
\end{table}
In the second strong baseline {\em Adversarial Discriminator F}, we use only $\hat{D}^{O \rightarrow F}$ and in 
{\em Adversarial Discriminator O} we use only $\hat{D}^{F \rightarrow O}$. These two baselines show the importance of channel-fusion. 

The results are shown in  Tab.~\ref{tbl:ablation}. 
 It is clear that {\em Adversarial Generator} achieves a very high accuracy: Comparing {\em Adversarial Generator} with all the methods in Tab.~\ref{tbl:UCSD}
(except our {\em Adversarial Discriminator}), it is the state-of-the-art approach.
Conversely, the overall accuracy of {\em Same-Channel Discriminator} drops significantly with respect to {\em Adversarial Discriminator} and is also clearly worse than {\em Adversarial Discriminator O}. This shows the importance of the cross-channel tasks. However, comparing {\em Same-Channel Discriminator} with  the values in 
Tab.~\ref{tbl:UCSD}, also this baseline outperforms or is very close to the best performing systems on this dataset, showing that the discriminator-based strategy can be highly effective even without cross-channel training. 

Finally, the worst performance was obtained by {\em Adversarial Discriminator F}, with values much worse than {\em Adversarial Discriminator O}. We believe this is due to the fact that {\em Adversarial Discriminator O} takes as input a real frame which contains  much more detailed information with respect to the optical-flow input of {\em Adversarial Discriminator F}. However, the fusion of these two detectors is crucial in boosting the performance of the proposed method {\em Adversarial Discriminator}.

It is also interesting to compare our {\em Adversarial Generator}
with the Convolutional 
Autoencoder proposed in \cite{DBLP:conf/cvpr/0003CNRD16},
being both based on the reconstruction error (see Sec.~\ref{sec:intro}).
The results of the Convolutional 
Autoencoder on the same dataset are:
$21.7\%$ and $90\%$ EER and AUC, respectively (Tab.~\ref{tbl:UCSD}), which are significantly 
worse than our  baseline based on GANs.


%
%

Finally, in the last row of  Tab.~\ref{tbl:ablation} we report the results recently published in \cite{2017icip}, where the authors adopted a strategy similar to the {\em Adversarial Generator} baseline above mentioned. The main difference between {\em GAN-CNN} \cite{2017icip} and {\em Adversarial Generator} is the use of an additional AlexNet-like CNN \cite{alexnet}, externally trained on ImageNet (and not fine-tuned) which takes as input both $F$ and the appearance generation produced by $G^{O \rightarrow F}(O)$ and computes a ``semantic'' difference  between the two images.
The accuracy results of {\em GAN-CNN} are basically on par with respect to the results obtained by the {\em Adversarial Discriminator} proposed in this paper. However, in {\em GAN-CNN} a fusion strategy needs to be implemented in order to take into account both the semantic-based and the pixel-level reconstruction errors, 
while the testing pipeline of
{\em Adversarial Discriminator} is very simple. Moreover, even if the training computation time of the two methods is the same, at testing time {\em Adversarial Discriminator} is much 
faster because  
$G^{O \rightarrow F}$, $G^{F \rightarrow O}$ and the semantic network are not used. 



\section{Qualitative results}
\label{sec:qualitative}

In this section we show some qualitative results of our generators 
$G^{F \rightarrow O}$ and $G^{O \rightarrow F}$ (Fig.~\ref{fig:qual2-O-F}) and some detection visualizations of the {\em Adversarial Discriminator} output. Fig.~\ref{fig:qual2-O-F} show that the generators are pretty 
good in generating normal scenes. However, high-resolution structures of the pedestrians are not accurately reproduced. This confirms that the  data distribution and the generative distribution do not completely overlap each other (similar results have been observed in many other previous work using GANs
\cite{DBLP:conf/nips/GoodfellowPMXWOCB14,DBLP:conf/nips/SalimansGZCRCC16,DBLP:journals/corr/RadfordMC15,Isola_2017_CVPR,nguyen2016ppgn}). On the other hand, abnormal objects or fast movements  are completely missing from the reconstructions:  the generators simply cannot reconstruct what they have never observed during training. This inability of the generators in reconstructing anomalies is directly exploited by both {\em Adversarial Generator} and {\em GAN-CNN} (Sec.~\ref{sec:ablation}) and 
intuitively  confirms our hypothesis that anomalies are treated as outliers of the data distribution 
(Sec.~\ref{sec:intro},\ref{sec:training}).

Fig.~\ref{fig:qual-det-succ} shows a few pixel-level 
detections of the {\em Adversarial Discriminator} in different
situations.
In Fig.~\ref{fig:qual-det-succ} the last column show some detection errors. Most of the errors (e.g., miss-detections) are due to the fact that the abnormal object is very small or partially occluded 
(e.g., the second bicycle)
and/or has a ``normal'' motion (i.e., the same speed of normally moving pedestrians in the scene). 
The other sample shows a false-positive example (the two side-by-side pedestrians in the bottom), which is probably due to the fact that their bodies are severely  truncated and the visible body parts appear to be larger than normal  due to perspective effects. 


\section{Conclusions}
\label{sec:Conclusions}

In this paper we presented a GAN-based approach for abnormality detection. We use the 
mutual supervisory information of our generator and  discriminator  networks 
in order to deal with the lack of supervised training data of a typical abnormality detection scenario. This strategy makes it possible to  
train  end-to-end anomaly detectors (our discriminators) using only
relatively 
small, weakly supervised training video sequences.
Differently from common  GAN-based approaches, developed for generation tasks, after training we directly use the discriminators as the final classifiers and we completely discard our generators. In order for this approach to be effective, we designed two non-trivial cross-channel generative tasks for training our networks.

As far as we know this is the first paper directly using a GAN-based training strategy for a discriminative task. Our results on the most common abnormality detection benchmarks show that the proposed approach sharply outperforms the previous state of the art.
Finally, we performed a detailed ablation analysis of the proposed method in order to show the contribution of each of the main components. Specifically, we compared the proposed approach with both  strong reconstruction-based baselines and same-channel encoding/decoding tasks, showing the overall accuracy and 
computational  advantages 
of the proposed method.

{\small
\bibliographystyle{ieee}
\bibliography{egbib}
}

\end{document}